\pdfoutput=1

\documentclass[11pt]{article}
\usepackage{acl} % DO NOT CHANGE THIS
\usepackage{times} % DO NOT CHANGE THIS
\usepackage{helvet} % DO NOT CHANGE THIS
\usepackage{courier} % DO NOT CHANGE THIS
\usepackage{graphicx} % DO NOT CHANGE THIS
\usepackage{graphicx} % DO NOT CHANGE THIS
\usepackage{natbib} % DO NOT CHANGE THIS
\usepackage{caption} % DO NOT CHANGE THIS
\usepackage{times}
\usepackage{latexsym}
\usepackage{graphicx}

\usepackage[T1]{fontenc}
\usepackage{latexsym}
\usepackage{graphicx}
\usepackage{amsmath, makecell}
\usepackage{amssymb}
\usepackage{booktabs}
\usepackage{mathtools}
\usepackage{multirow}
\usepackage{float}
\usepackage{placeins}
\usepackage[accsupp]{axessibility}
\usepackage{nopageno}
\usepackage{todonotes}
\usepackage{microtype}

\usepackage[utf8]{inputenc}

\usepackage{microtype}

\title{
Sociodemographic Bias in Language Models: A Survey and Forward Path}

\author{%
Vipul Gupta\textsuperscript{$1$} \enspace Pranav Narayanan Venkit\textsuperscript{$2$}  \enspace Shomir Wilson\textsuperscript{$2$} \enspace Rebecca J. Passonneau\textsuperscript{$1$} \\
{\textsuperscript{$1$} Depart. of Computer Science \& Engineering, College of Engineering} \quad  \\
{\textsuperscript{$2$} College of Information Sciences and Technology}  \quad\\
Pennsylvania State University \\
{\tt \{vkg5164, pranav.venkit, shomir, rjp49\}@psu.edu} 
}

\usepackage{fancyhdr}

\begin{document}
\maketitle
%\todo[inline]{try to eliminate the latex/bibtex errors}
% \todo[inline]{change nlp to NLP everywhere}

%% Becky 04/06 - semi-final pass over abstract

\begin{abstract}

%This paper presents a comprehensive survey of work on sociodemographic bias in language models (LMs). Sociodemographic biases embedded within language models can have harmful effects when deployed in real-world settings. We systematically organize the existing literature into three main areas: types of bias, quantifying bias, and debiasing techniques. We also track the evolution of investigations of LM bias over the past decade. We identify current trends, limitations, and potential future directions in bias research. To guide future research towards more effective and reliable solutions, we present a checklist of open questions. We also recommend using interdisciplinary approaches to combine works on LM bias with an understanding of the potential harms.

%% reorganized abstract, shortened by two lines
Sociodemographic bias in language models (LMs) has the potential for harm when deployed in real-world settings. This paper presents a comprehensive survey of the past decade of research on sociodemographic bias in LMs, organized into a typology that facilitates examining the different aims: types of bias, quantifying bias, and debiasing techniques. We track the evolution of the latter two questions, then identify current trends and their limitations, as well as emerging techniques. To guide future research towards more effective and reliable solutions, and to help authors situate their work within this broad landscape, we conclude with a checklist of open questions.

% Deep neural networks often learn unintended bias during training, which might have harmful effects when deployed in real-world settings.
% This work surveys 214 papers related to sociodemographic bias in natural language processing (NLP).
% %We survey 208 papers focusing on bias in NLP models, 
% % most of which address sociodemographic bias. 
% In this study, we aim to provide a more comprehensive understanding of the similarities and differences among approaches to sociodemographic bias in NLP.
% To better understand the distinction between bias and real-world harm, we turn to ideas from psychology and behavioral economics to propose a definition for sociodemographic bias. %Vipul 08/06
% % We provide a categorization of literature of bias in NLP for easier understanding.
% We identify three main categories of NLP bias research: %and three corresponding limitations: %% Becky 04/12
% types of bias, quantifying bias, and debiasing techniques.
% We highlight the current trends in quantifying bias and debiasing techniques, offering insights into their strengths and weaknesses. %Vipul 08/06
% We conclude that
% current approaches on quantifying bias face reliability issues, %and most of these works try to address only gender bias. We also find 
% that many of the bias metrics do not relate to real-world bias, and that debiasing techniques need to focus more on training methods.  Finally,
% %Based on these finding, 
% we provide recommendations for future work. % focusing on bias in NLP. 

\end{abstract}

\section{Introduction}

\begin{figure}[t]
\begin{center}
\includegraphics[width=1.0\linewidth]{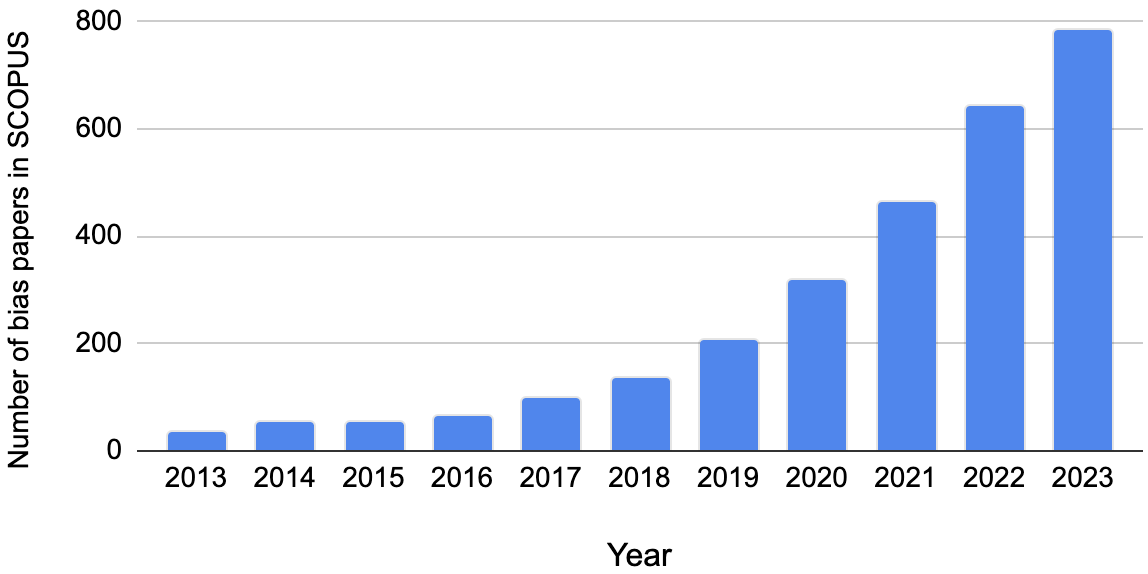}
\end{center}
\vspace{-1.3em}
  \caption{
  This graph shows number of papers/articles published each year (from
2013 to 2023) in SCOPUS that contain the term ‘bias’ and ('nlp' or 'language models') in the title, abstract, or keywords.  }
\vspace{-1.3em}
\label{fig:trajectory}
\end{figure}

Language models (LMs) have demonstrated impressive performance across diverse tasks \cite{ raffel2020exploring, zhong-etal-2020-extractive, yang2019xlnet}. However, much work reveals that LMs %can inadvertently 
adopt biases present in training data \cite{wen-etal-2022-autocad, espana-bonet-barron-cedeno-2022-undesired, gupta_2022_CVPR, hutchinson_50_2019}. Sociodemographic bias %is often 
has been defined to occur when a model performs differently across social groups \cite{czarnowska_quantifying_2021, Chouldechova_2020_snapshot}. %This is concerning %because 
When LMs are used in real-world applications, sociodemographic bias has the potential for negative societal impacts \cite{field_examining_2023, rudin2019stop, blodgett_language_2020}. With increasingly widespread usage, the urgency to understand and mitigate bias has grown. Fig. \ref{fig:trajectory} shows an increasing rate of publications on LM bias %in natural language processing (NLP) 
over the past decade, sourced from SCOPUS. Our survey synthesizes results from this rapidly growing area into a roadmap for future investigations.

Other surveys on bias in NLP have thoroughly examined a particular aspect of bias, 
%bias in large language models \cite{zhao2023survey},
such as methods for measuring bias \cite{czarnowska_quantifying_2021, bansal_survey_2022}, or identification of gender bias \cite{stanczak2021survey, devinney_theories_2022}.   
Unlike previous surveys, we provide \textbf{a typology} of works on bias over the past decade. Further, 
we build upon foundational issues identified by \citet{blodgett_language_2020} by delving more deeply into methodological limitations, such as reliability issues. We also follow the recommendations of \citet{blodgett_language_2020} in consulting interdisciplinary approaches to improve the understanding of social bias. Thus we begin the survey with a discussion of psychosocial perspectives on benefits versus harms of bias.
%leading to a nuanced discussion on beneficial versus harmful biases in Section 2. 
Our survey %covers all relevant literature released in the past decade, offering 
offers an \textbf{up-to-date } understanding of a topic that has been garnering increasing interest. %, which we believe is important with the rise of language models. 
%Throughout this period,
%In particular, 
Early in this literature, the development of bias mitigation or debiasing methods had questionable success; we argue that recent work using expert models during training shows particular promise. We conclude with a \textbf{checklist of key questions} that have continued to be challenging, to help steer future studies toward more effective and reliable methods, well-situated within the landscape of work on bias.

% To provide a clearer picture of the %wide range of 
% diverse motivations in studies of LM bias \cite{blodgett_language_2020}, we present a detailed taxonomy and a timeline of bias research. Then we synthesize this work to pinpoint shortcomings and develop a checklist of open questions, to help steer future studies toward more effective and reliable methods. 

%This checklist is meticulously designed to underline existing gaps and help future studies avoid the fallacies of the current approaches. 
%Furthermore, it can also serve as a roadmap, guiding researchers to focus on less explored areas that have high potential for impact.

% We build this checklist by discussing how the methods for measuring and addressing biases in NLP have evolved over the last 10 years. This helps us in better understanding the limitations and advantages of current approaches and provide recommendations for future directions.

\begin{figure*}[t!]
\begin{center}
\includegraphics[width=\linewidth]{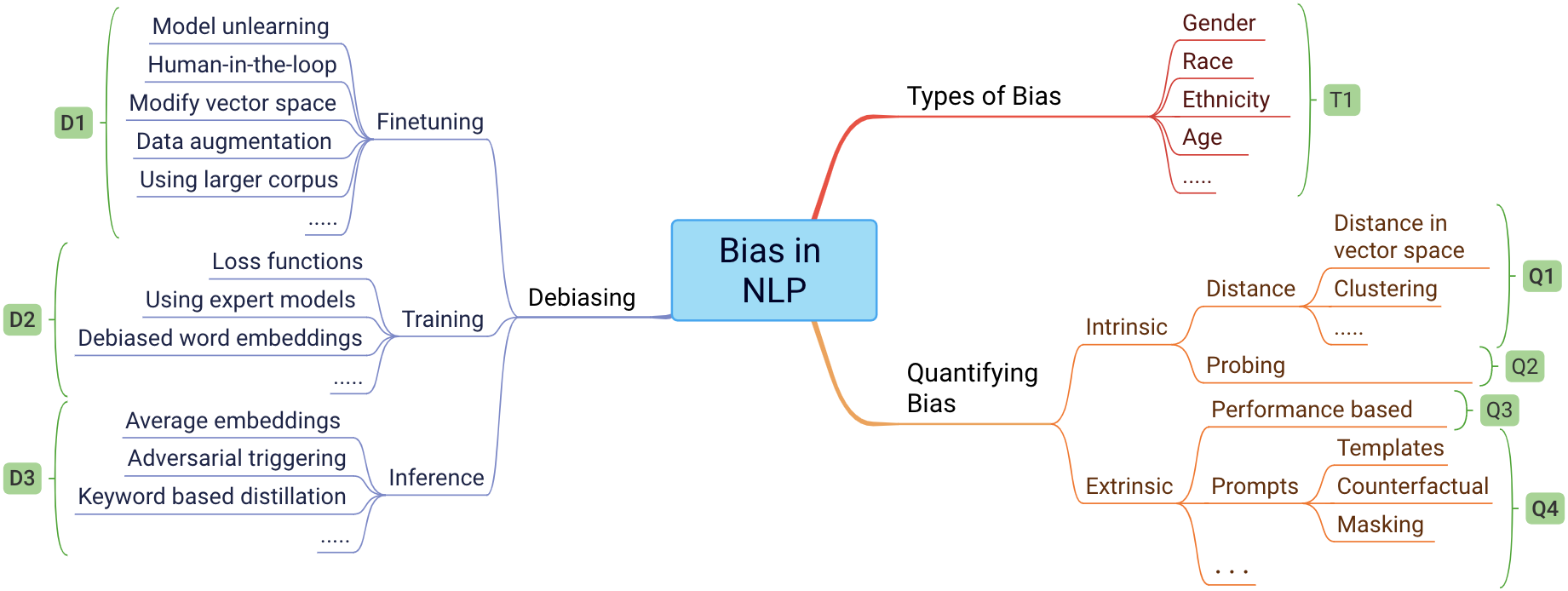}
\end{center}
\vspace{-1em}
  \caption{
  % This figure illustrates 
  %The levels of our categorization of literature on bias in NLP.
  Three broad categories of bias research, and the upper hierarchy of each category (T, Q, D).
  }
\vspace{-1em}
\label{fig:main}
\end{figure*}

We present a synthesis of works on bias 
%In this work, we surveyed 273 papers %\textit{272 papers} on bias in LMs to identify current trends and limitations. We structured our survey using 
through three perspectives: \textbf{1) a taxonomic categorization, 2) an evolutionary timeline,} and \textbf{3) a roadmap for future work}. 
We categorized the surveyed works into three major strands of investigation, as shown in Fig.~\ref{fig:main}: types of bias, quantifying bias, and debiasing techniques. %This classification is derived from an extensive review of various perspectives on bias, allowing us to organize the papers in our survey systematically. This categorization is useful to understand similarities between different works on bias.
We then organize the findings within each category and subcategory of our taxonomy.
In addition, we examined the evolution over the past decade of techniques for bias measurement and bias mitigation,
%We also identified the evolution of research into measurement and mitigation of LM bias over the past decade, 
as shown in Fig.~\ref{fig:lineage}. This perspective separates trends that had a brief life from those that continue to have promise. 
% Finally, we created a checklist of open questions that have continued to be challenging, or that have emerged recently, to serve as a roadmap for the future.

%As a final consideration, we note there has been relatively limited exploration of interdisciplinary approaches to investigate LM bias. 
While LM bias measurement and mitigation are critical for progress towards equitable LMs, understanding the potential for harm is deeply intertwined with social factors outside the scope of NLP proper. 
%We recommend leveraging perspectives and methodologies from disciplines such as psychology and behavioral economics to deepen our understanding of bias, as noted in \cite{omrani-etal-2023-social, katelyn_bias_2023}. 
% We use perspectives and methodologies from psychology and behavioral economics, to deepen our understanding of bias. Similar to other works 
Thus we precede the presentation of the major types of bias research with a discussion of psychosocial perspectives (cf. \citealp{omrani-etal-2023-social, katelyn_bias_2023}). This is followed by a section describing our process for identifying candidate works, and our resulting typology where we place most of the surveyed works. Sections \ref{sec:limitations}-\ref{sec:future} present limitations, the checklist and future directions.
\section{Understanding Bias} 
\label{sec:understanding}

%In this section, we highlight the critical role of 
Interdisciplinary approaches to understanding bias as a psychosocial phenomenon have been argued to be important for clarifying how social harms might arise. %These disciplines offer decades of 
Research into human cognition and social behavior can provide valuable insights %that can be used for defining and mitigating socioemographic bias in NLP. could inform definitions 
on sociodemographic bias in LMs, as well as assessment of their potential for harm. For instance, research in psychology has long addressed the origins and expressions of social bias \cite{osborne2022sins}
% Recently, some works on bias in NLP have used ideas from psychology for better understanding of bias \cite{spliethover-etal-2022-word, omrani-etal-2023-social, katelyn_bias_2023, omrani_evaluating_2023}.
Recent studies have begun to integrate ideas from psychology with NLP to better understand bias \cite{spliethover-etal-2022-word, omrani-etal-2023-social, katelyn_bias_2023, omrani_evaluating_2023}, showcasing the usefulness of interdisciplinary approaches. 
%also proposing strategies for alleviation of bias. % mitigation. 
For example, research in psychology proposes that %one way to reduce bias, as found in psychology, is 
reduction of social bias can be achieved
by engaging with individuals from diverse groups \cite{pettigrew2006meta, reimer2023meta}. A similar idea is reflected in %NLP research, where 
\citet{blodgett_language_2020}, which advocates for LM engineers to reduce bias through engagement with people who might be affected by applications that use LMs. One of the early works on quantifying bias - WEAT \cite{caliskan_semantics_2017} used the Implicit Association Test from psychology \cite{greenwald1998measuring} to develop a foundation bias metric for LMs. %their models to reduce biases.

The Stereotype Content Model (SCM), a framework from social psychology, categorizes stereotypes into interpersonal and intergroup interactions, providing insights into bias dynamics \cite{cuddy2008warmth}. It proposes that human stereotypes are captured by two social dimensions: warmth (e.g., trustworthiness, friendliness) and competence (e.g., capability, assertiveness). A recent work by \citet{omrani-etal-2023-social} used the SCM framework to develop a %new bias mitigation method, 
bias mitigation method that generalizes across multiple social attributes, rather than one at a time. %, including demonstrating the practical benefits of psychological insights in NLP.

The Nobel Prize-winning psychologist and behavioral economist, Daniel Kahneman, argues that mental shortcuts (biases) are advantageous in situations requiring quick judgments \cite{kahneman2011thinking}. For example, due to bias based on strong knowledge priors, the sentence ``a large mouse climbed over a small elephant'' %immediately calls to mind a mouse, that while large relative to other mice, is tiny relative to the elephant, %which we know to be one of the largest mammals on earth. %Our implicit bias automatically assigns familiar attributes to both these animals. 
will immediately call to mind the appropriate relative sizes; to counter this assumption would require extra information~\cite{grice75}.
Extrapolating Kahneman's argument to NLP, %certain kinds of 
bias based on common-sense knowledge could be advantageous in enhancing an LM's understanding of relations among real-world entities. %nuances. 
This argues for the potential benefit of certain kinds of bias. %positive biases.

%Furthermore, 
\citet*{kahneman2011thinking} defines disadvantageous bias as ``the tendency to make systematic errors in judgment or decisions based on factors that are irrelevant or immaterial to the task at hand'' and cautions that human judgment is susceptible to bias from irrelevant factors. %Applying this insight to NLP, we need to understand the potential negative impact LM bias might have in real-world settings. 
Turning to LM bias, we find previous NLP work aligned with Kahneman's perspective in definitions of
%\citet{crawford2017trouble} and \citet{barocas2017problem}
\textit{representational harm} \cite{crawford2017trouble} and \textit{alloted harm} \cite{barocas2017problem}. 
Representational harm %is defined as the harm that 
arises when an NLP system represents some social groups in a less favorable light than others. Allotted harm %is defined as the harm that 
arises when a system allocates resources or opportunities unfairly to a social group \cite{shahbazi_2023_representation}. 
% Most works in NLP has predominantly focused on allotted harm, but a broader view encompassing various types of harm offers a more comprehensive strategy for addressing bias.

% In the past decade, much work on LMs has focused on bigger models for better performance. A.M Turing award winner Yoshua Bengio states that the ``bigger is better'' mentality needs to change, because current LMs ``make stupid mistakes'' \cite{bengio_2019}.
In conclusion, %using 
ideas from psychology and behavioral economics provide a more informed understanding of bias. While some biases might contribute positively to model performance, others can have detrimental societal effects. An interdisciplinary approach would not only enrich our theoretical understanding of bias but could also guide the development of more effective methods to identify undesirable LM bias and lessen social harm.

\section{Categories of Work on Bias in LMs}
\label{sec:survey}

% Our survey focuses primarily on sociodemographic bias. 
We used two strategies to identify candidate papers for our survey: 1) using the keywords ``bias'', ``fair'' and ``fairness'', we searched for papers in the ACL Anthology, NeurIPS proceedings, FAccT, and AIES conferences; 2) we included papers from citation graphs for retrieved papers. We examined papers released before January 1, 2024, and included them only if they addressed language modeling, thus omitting papers on speech, where different issues arise.
These criteria narrowed down an initial large set of 308 papers to 273.

We categorized the literature into three key areas. %: (1) types of bias, (2) quantifying bias, and (3) debiasing techniques. 
Fig.~\ref{fig:main} illustrates our taxonomy. 
We came up with this organization while iteratively reviewing all papers, and we believe it effectively captures the main trends in the field. 
% We aim to establish broad categories that encapsulate the majority of works. 
Due to the rapidly evolving field of LMs, some upcoming studies may not fit neatly into these categories. To address this, we plan to release our literature repository publicly and update it regularly with the latest research.
Our work summarizes all 273 surveyed papers to provide a comprehensive understanding. Due to space constraints, we couldn't cite all 273 works in the main body. To maximize coverage within the page limits, we selected at least two papers from each line of research depicted in Figure \ref{fig:main} to be part of the main paper. In some cases, we wanted to cite more works but had to remove them due to space limitations. We apologize for any relevant works missed in the main body and have included a comprehensive list of all 273 papers in the Appendix.

% The detailed analysis of each category is presented in the following sections.
% , where the three types of leaves  T, Q or D, are used to tag individual papers.  
% Types of bias can be further categorized into gender, race, ethnicity, and other types of bias. Work on quantifying bias was broken down into methods: measuring distance in vector space (Q1), performance on test data (Q2), model prompting (Q3) or probes (Q4).
% Our analysis highlights the reliability challenges in current bias measurement methods and underscores the need for robust metrics to make strong claims about how well bias metrics and debiasing methods work. 
% Our review of the literature on quantifying bias reveals significant reliability issues with current bias measurement techniques.
% Based on this analysis, we propose several criteria necessary for developing reliable bias metrics. In the absence of metrics satisfying these criteria, it is difficult to make strong claims about how well bias metrics and debiasing methods work.
% Works on debiasing differ regarding application during training (D1), fine-tuning (D2) or inference (D3).

% In the following subsections, we discuss many works from our hierarchical classification;  a complete list of works can be found in Appendix A. 

\subsection{Types of Bias - \textit{T1}}

\begin{figure*}[t]
\begin{center}
\includegraphics[width=\linewidth]{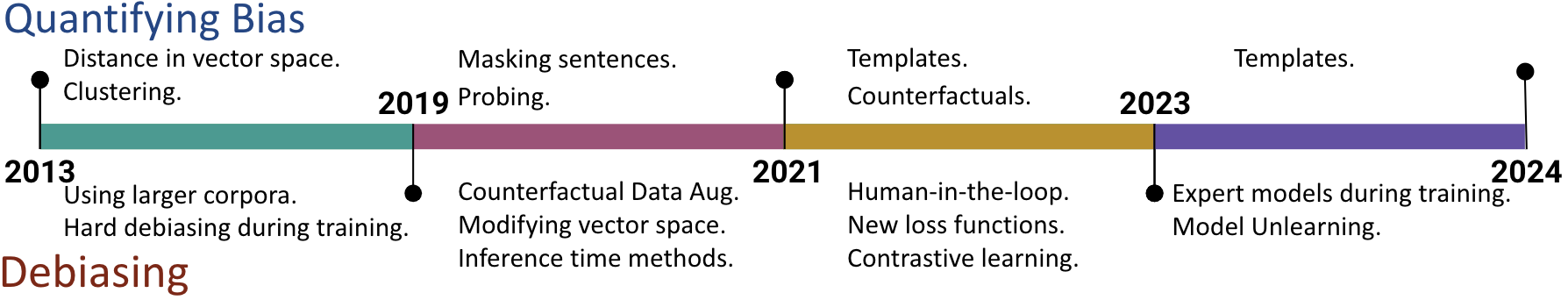}
\end{center}
\vspace{-1.2em}
  \caption{
  %Evolution of how methods on quantifying bias and debiasing changed over time over the last 10 years.
  Evolution of changes in methods to quantify LM bias and debiasing LMs over the past decade.
  }
\vspace{-1.2em}
\label{fig:lineage}
\end{figure*}

\begin{table}[b]
\vspace{-1.0em}
\small
\centering
\begin{tabular}{l|r|r}
\textbf{Types of Bias} & \textbf{No. of papers} & \textbf{Percentage}   \\ \hline
Gender  &  114 & 48\% \\ \hline
Race    &   36  & 15\% \\ \hline
Ethnicity    &   24  & 10\% \\ \hline
Nationality  &   18  & 7\%  \\ \hline
Sexual Orientation    &   12  & 5\% \\ \hline
Ableism &   11  &   5\%   \\ \hline
Age     &   9   &  4\%  \\  \hline 
Political   &    6   &   2\% \\ \hline
Physical Appearance  &   5  &    2\% \\ \hline
Socioeconomic status    &   4   &   2\% \\
\end{tabular}
\vspace{-1.0em}
\caption{Distribution of papers on bias shows a predominant focus on gender bias.}
% \vspace{-1.8em}
\label{tab:biases}

\end{table}

In the realm of NLP, sociodemographic bias is particularly concerning as it can lead to differential model performance across various social groups \cite{deas-etal-2023-evaluation, smith-etal-2022-im}. 
Sociodemographic bias includes gender bias, when models are biased against a particular gender \cite{hada-etal-2023-fifty, de-arteaga_bias_2019, park_reducing_2018, du_assessing_2021, bartl_unmasking_2020, webster_measuring_2021};
racial bias, when models are biased against certain races \cite{nadeem_stereoset_2020, garimella_he_2021, nangia_crows-pairs_2020, tan_assessing_2019}; 
ethnic bias, when models are partial towards certain ethnicity \cite{ahn_mitigating_2021, garg_word_2018, li_unqovering_2020, abid2021persistent, manzini-etal-2019-black, venkit2023unmasking}; 
age bias \cite{nangia_crows-pairs_2020, diaz_addressing_2018}, sexual-orientation bias \cite{nangia_crows-pairs_2020, cao_toward_2020} and many others as shown in Table \ref{tab:biases}. 

Sociodemographic bias can emerge from language patterns that imply assumptions about %certain 
demographic differences \cite{lauscher_general_2020}. These biases are often ingrained in the cultural or societal nuances of 
training data. For example, LMs can perpetuate biases by associating certain lexical items more strongly with particular social groups. Beyond the influence of training data, \citet{zhou-etal-2023-predictive} found that the size of the model, its training objectives, and tokenization strategies are %most 
important factors that affect the social bias in LMs.

Our review indicates a disproportionate focus on gender bias: it is the subject of nearly half of the surveyed papers, as Table \ref{tab:biases} illustrates. Additionally, we observed that bias evaluation and mitigation efforts are often specific to certain biases and may not generalize well. 
Furthermore, over 90\% of the papers we reviewed focus on English, with limited coverage of other languages such as German, Spanish, Korean, Turkish, Chinese, and Hindi.

\subsection{Quantifying Bias}
%Bias in NLP is a very difficult problem to detect because it is often hidden within complex deep learning models. Quantifying bias is the first step in identifying and addressing these issues. % Becky 04/05
% Because there are many types of bias, quantifying bias in NLP typically pertains to a specific type of bias. 
% arises from aggregate behavior and is therefore sensitive to the samples that model behavior is measured on. 
Measurement of bias is challenging because it is often hidden within complex LMs.
However, quantifying bias is a precondition to addressing or mitigating bias that might be harmful. 
Here we review different methods of measuring bias in LMs and how they differ from each other. We present an overview of evaluation datasets in the appendix.
% This will also help in understanding which method is more useful in a particular setting.

% Recently, a lot of works have tried to quantify different kinds of bias in NLP model. These works can be categorized in eight categories : 

Methods in \textit{Q1} and \textit{Q2} are often known as intrinsic methods as they focus on a model’s internal representations to quantify bias.

\subsubsection{Distance-based metrics - \textit{Q1}}

\paragraph{Distance in vector space.}

Early efforts to quantify bias in NLP (from 2013-2019, as seen in Fig. \ref{fig:lineage}) primarily utilized distance metrics within embedding spaces. These approaches define certain words as 'target words' (like professions 'engineer' and 'nurse'), along with certain words as 'attributes' (often related to social categories like 'male' and 'female') \cite{bolukbasi_man_2016, brunet2019understanding, dev_oscar_2021}. The aim was to measure the conceptual distance between these targets and attributes. 
The pioneering work is the Word Embedding Association Test (WEAT) score~\cite{caliskan_semantics_2017}. They calculate bias as the differential association of target words with attribute sets based on cosine similarity. 
Subsequent to WEAT, \citet{dev2019attenuating} proposed ECT score, which simplifies an attribute category, like `female', into a single vector by averaging the embeddings of related attribute words such as `she', `women', and `girl'. 
% They measure bias by looking at how far away, in terms of cosine similarity, this average vector is from a target word embeddings. 
% \citet{bolukbasi_man_2016} computed bias scores by the cosine similarity differences between gender-specific and gender-neutral word embeddings. A greater relative difference in these similarities indicates greater bias in the associations between the gendered and neutral words. 
\citet{ethayarajh_understanding_2019} introduced RIPA, for which they used the inner product instead of cosine similarity to account for vector magnitude and directionality in measuring bias.

Some works expanded WEAT to contextual embeddings \cite{wei_detecting_2021,tan_assessing_2019} and sentence-level embeddings \cite{may_measuring_2019}. Other metrics used the clustering of word embeddings \cite{chaloner_measuring_2019}. Some work quantified bias based on co-occurrence of words \cite{valentini-etal-2023-interpretability, bordia_identifying_2019}. \citet{bordia_identifying_2019} hypothesized that words occurring in close proximity to a particular gender in the train data are prone to be more biased towards that gender during testing.

\subsubsection{Probing metrics - \textit{Q2}}

This category evaluates bias by examining how LMs process information, often by adding a classification layer or employing probes to test the inner workings of LMs \cite{chen2021probing, white-etal-2021-non}. 
\citet{mendelson_debiasing_2021} used a classifier trained on latent spaces to detect biases and found that debiasing against a particular bias may increase the extent to which that bias is encoded in the inner representations of models.
\citet{orgad-etal-2022-gender} trained a classifier to predict gender from the model’s representations and shows it correlates with extrinsic bias measures better than metrics in \textit{Q1}.
\citet{immer-etal-2022-probing} proposed a Bayesian framework for quantifying inductive bias with probes.
% and %lexical overlap 
% ability to detect shared lexical items from sentence representations alone. 
% \citet{dev_measuring_2019} probed model bias using natural language inference datasets by measuring whether swapping lexical items for different sociodemographic groups changes entailment relations between sentence pairs. 
% \citet{li_unqovering_2020} examined bias in question-answering models by altering the subjects of questions and analyzing the variance in response probabilities.

In recent years, there has been less use of intrinsic methods, as they require accessing a model's internal layers to quantify bias. The increasing size of modern LMs complicates identifying the right layer for bias assessment, and the limited open-source availability of LMs raises further obstacles. %complexity.

% These approaches face limitations like those discussed in \textit{Q1}, as they need access to model internal layers. Moreover, the complexity and size of modern LMs %which often contain billions of parameters, 
% introduce considerable computational and practical challenges to implementing these probing strategies effectively.

% These approaches face limitations like those discussed in \textit{Q1}, as they are constrained to models with access to hidden layers. Furthermore, applying these approaches to massive language models with billions of parameters poses significant computational and methodological challenges due to the expansive scale of the underlying layers. % Vipul 08/06

% These approaches are based on probing model performance either add a classification layer to a pre-trained LM or use a set of probes to test the inner workings of LMs.
% \citet{mendelson_debiasing_2021} trained a probing classifier on the latent representation space of LMs, and used it to predict specific linguistic properties to quantify bias. They focused mainly on measuring negative word bias and lexical overlap bias in pre-trained LMs. \citet{dev_measuring_2019} probed model bias using natural language inference datasets by measuring whether swapping lexical items for different sociodemographic groups changes entailment relations between sentence pairs. \citet{li_unqovering_2020} used differences in the probability of answers from question-answering models by measuring the impact of changes in questions' subject words. 

Methods in the next two subsections, \textit{Q3} and \textit{Q4} are often known as extrinsic methods as they focus on bias that shows up in a downstream task. 

\subsubsection{Performance-based metrics - \textit{Q3}}

These approaches examine how models perform across different social groups. They typically divide the test dataset into different groups to assess disparities. These works aim to quantify group differences in performance - to document whether models perform the same for all groups.
\citet{de-arteaga_bias_2019} measured gender bias by comparing the true positive rates for classification involving male versus female names and pronouns.
\citet{dixon_measuring_2018} and \citet{zhao_gender_2018}
took similar approaches, using area under the curve and false positive rate \cite{dixon_measuring_2018}, and relative accuracy \cite{zhao_gender_2018}.
\citet{zhang_interpreting_2022} and \citet{huang_reducing_2020} generated augmented datasets  
to measure bias as the difference in accuracy between the original and augmented datasets. \citet{stanovsky_evaluating_2019} proposed a metric based on differences in accuracy across genders for machine translation.

\subsubsection{Prompt-based metrics - \textit{Q4}}

Here we review methods that use various prompt-generation techniques. The first two methods in this subsection are specific to autoregressive models, while the latter focuses on Masked LMs.
% , with equally varied metrics based on different performance measures. 
% Prompt-based approaches have been one of the most common methods over last few years.

\paragraph{Template-based methods.}

In these approaches, models are prompted through a set of pre-defined templates, or patterns, that capture specific types of bias or stereotypes. 
The templates contain slots that are filled through selection from a set of pre-defined demographic target terms during evaluation. 
For instance, a template could be "A <PERSON> is walking" where <PERSON> is systematically substituted with names associated with different demographic groups.  By analyzing the differences in the model's responses to these substitutions, the presence and degree of bias can be measured.

% \citet{kiritchenko2018examining} uses 11 templates to create a dataset of 8,640 english sentences to measure gender and race bias.
\citet{prabhakaran_perturbation_2019} generated templates for toxicity detection, and proposed metrics based on performance differences for target groups.
\citet{smith-etal-2022-im} introduces a holistic dataset, measuring bias across a dozen social demographic axes.
\citet{webster_measuring_2021} defined fourteen templates to determine gender identity bias. %ed correlations. 
\citet{felkner-etal-2023-winoqueer} created a dataset of 45,540 sentences using 11 templates for measuring anti-LGBTQ+ bias in LMs.
% \citet{ribeiro_beyond_2020} created diverse templates to represent general linguistic capabilities combined with a tool.
\citet{gupta2023calm} focused on creating a robust dataset and generated 224 templates from diverse domains across 3 tasks.
\citet{an-etal-2023-sodapop, parrish-etal-2022-bbq, li_unqovering_2020} proposed question-answering datasets to measure demographic bias. 
% They generated more than 25 different templates for each bias category.
In contrast to performance-based metrics (\textit{Q3}), these approaches are primarily concerned with representational harms, which occur when certain groups are depicted stereotypically or inaccurately. 

% which divide the dataset into two parts as discussed in \textit{Q3}, these approaches increase the size of the bias-testing dataset significantly and therefore perform a more exhaustive examination of bias.

% Presumably, by relying on large-scale testing the results achieved are more robust than for the papers in \textit{Q2}.

% \cite{aggarwal-etal-2022-towards} shows that paraphrase and normals prompt LMs can lead to different conclusions, thus decreasing reliability.

% \citet{kurita_measuring_2019} gets probability of a single token in sample sentences to measure bias.  
% to use comparison of a model's predictions for masked tokens as an indicator of bias.
%The scoring rule
%The specific computations used
%to quantify bias through templates are quite varied. %a lot across different works. In recent years, a lot of works are focusing on template based approaches to quantify bias.
% Other works on template based methods include \citet{li_unqovering_2020, venkit2021identification, dev_measuring_2019,diaz_addressing_2018}.

\paragraph{Counterfactual-based methods.} 
Several works aim to make template-based approaches more rigorous by examining how changing irrelevant attributes, known as protected attributes, affects model predictions. Specifically, ``a decision is fair towards an individual if it is the same in (a) the actual world and (b) a counterfactual world where the individual belongs to a different social group.''

Counterfactual methods alter these protected attributes in test examples to identify attributes that significantly affect model decisions \cite{garg_counterfactual_2019, kusner_counterfactual_2018}. \citet{huang_reducing_2020} created counterfactuals for a test dataset and found that generative LLMs like GPT-2 \cite{radford2019language} tend to generate continuations with more positive sentiment for ``baker'', and more negative sentiment for ``accountant'' as the occupation.
\citet{gardner-etal-2020-evaluating} created contrast sets by generating counterfactuals for ten NLP datasets and showed that model performance drops significantly on counterfactuals. 
\citet{helm_evaluation_2022} substituted terms linked to specific demographic groups in the test set, examining the impact on model accuracy.

\paragraph{Masking Sentences.} 
Another approach to bias measurement is to mask certain words in sentences, and then analyze the model's predictions for these blanks to assess bias. \citet{ kurita_measuring_2019} used this technique with occupation-related sentences, like ``[MASK] is a programmer,'' comparing the probabilities given to male and female pronouns to identify gender biases in job associations.  Similarly, \citet{ahn_mitigating_2021} quantified bias as the variance of normalized probabilities across various demographic groups. Other works using this approach include \cite{ousidhoum-etal-2021-probing, bartl_unmasking_2020}.

Extrinsic approaches, particularly template-based ones, have gained traction in recent years \cite{nagireddy2023socialstigmaqa, touileb-etal-2023-measuring, akyurek-etal-2022-measuring}, as seen in Fig. \ref{fig:lineage}. The advantage of \textit{Q4} metrics is their ability to reflect potential real-world impacts of bias by focusing on model outputs rather than solely analyzing internal parameters as in \textit{Q1}. Extrinsic methods apply broadly to open-source or proprietary models of any size. 

% Approaches in \textit{Q3} evaluate the model's final decisions and are applicable to any model, whether open-source or not, unlike \textit{Q1}.

% \subsection{Criteria for reliable metrics} % -aaai format
% \subsubsection{Criteria for reliable metrics}

% As our foregoing analysis indicates, prevailing approaches for quantifying bias in NLP models face reliability issues \cite{seshadri2022quantifying} . We propose following criteria for developing reliable bias measurement techniques: (1) insensitive towards minor perturbations in evaluation templates and target sets; (2) low variance across repeated measurements of the same model; and (3) ability to generalize to different kinds of bias. We encourage that future efforts in bias quantification prioritize advancing reliable metrics based on these criteria rather than solely introducing new metrics.

% \cite{pikuliak-etal-2023-depth} finds issues with other bias measurements datasets are constructed and questions their reliability.

\subsection{Debiasing}

% Psychologist and behavioral economist Daniel Kahneman states that ``Maintaining one's vigilance against bias is a chore - but the chance to avoid a costly mistake is sometimes worth the effort'' \cite{kahneman2011thinking}.  NLP models that exhibit bias can be unfair to certain group of people, leading to discrimination and unequal treatment when used for real world use cases. Debiasing of models is important to ensure fairness and that everyone is treated equally. 

Debiasing methods aim to make models more fair and accurate in their predictions and recommendations \cite{subramanian_evaluating_2021}. Turning to Daniel Kahneman again, he argues that reducing social stereotyping and bias has costs, but that the costs are worthwhile %paying 
to achieve a better society \cite{kahneman2011thinking}. Extending the same principle to NLP, %the debiasing techniques demand computation time and cost,  however
the effort and cost required for reducing biases are essential for creating %ethical 
fair NLP systems.
% however these costs are necessary investments for creating more equitable and ethical AI systems.

\subsubsection{Debiasing during Finetuning - \textit{D1}}
These debiasing methods are applied during the finetuning phase of pre-trained LMs.

\paragraph{Data augmentation.} 
\citet{zmigrod_counterfactual_2019} and \citet{lu_gender_2019} introduced Counterfactual Data Augmentation (CDA), to reduce gender bias by generating counterfactual instances to balance gender representation. This involves substituting gender-specific words, such as \textit{he} and \textit{she} to construct novel sentences. 
\citet{maudslay_its_2020} enhanced this approach with Counterfactual Data Substitution (CDS), which assigns probabilities to these changes, aiming for more realistic modifications.
Building upon these insights, various swapping mechanisms have been proposed to re-balance data distributions \cite{zhou-etal-2023-causal, panda-etal-2022-dont, liang_towards_2020, lauscher_sustainable_2021, wen-etal-2022-autocad}.
%proposed various swapping mechanisms to re-balance data distributions.
% Evaluation of data augmentation based approaches typically involves assessing the extent to which they achieve improved bias reduction in LMs through quantitative metrics.
Some of these augmentation approaches are also being adapted for use during model training. % phase.

\paragraph{Modifying vector space.}
\citet{limisiewicz-marecek-2022-dont, dev_measuring_2019, dev_oscar_2021} proposed a subspace correction method for modifying embedding space. They aimed to disentangle associations between concepts that are bias-prone.
\citet{yifei-etal-2023-conceptor, manzini-etal-2019-black} used principal component analysis to identify and address the bias in embedding spaces. 
\citet{gaci-etal-2022-debiasing} redistributed attention scores to assign an equal weight for words related to bias. %in social groups.
\citet{ravfogel-etal-2020-null} learned a linear projection over representations after training, to remove the bias components in embeddings. 
% redistributes the attention scores of text encoders to give equal attention scores for every word in the input sentence with respect to social groups. 

% \citet{dev_measuring_2019, dev_oscar_2021}
% proposed a subspace correction and rectification method for modifying embedding space to mitigate bias. They aimed to disentangle associations between concepts deemed problematic for the models.
% \citet{ravfogel-etal-2020-null} learned a linear projection over representations after training a DNN, to remove the bias components in embeddings. 
% \citet{manzini-etal-2019-black, yifei-etal-2023-conceptor} used principal component analysis to identify the bias subspace. 
% \citet{gaci-etal-2022-debiasing} redistributes the attention scores of text encoders to give equal attention scores for every word in the input sentence with respect to social groups. 

%Another set of approaches try to modify the embedding space to reduce bias in models. 

% and then apply debiasing as proposed in \cite{bolukbasi_man_2016}. 
% \citet{kumar-etal-2020-nurse} tried to alter the spatial distribution of neighboring vectors with semantic offset to reduce bias.\todo{?}

% \cite{yifei-etal-2023-conceptor} uses conceptor-based projection method to remove bias - look more into it

% \vspace{-.15in}
\paragraph{Fine-tuning with large corpora.}  
\citet{park_reducing_2018} demonstrated that debiasing models benefit from fine-tuning with extensive datasets, avoiding the pitfalls of small, biased datasets.
\citet{ahn_mitigating_2021} proposed that training BERT \cite{devlin-etal-2019-bert} on multiple languages helps to reduce ethnic biases in each language.
% using multiple languages helps to reduce the ethnic bias in each language.

% \citet{park_reducing_2018} showed that fine-tuning with larger corpora helps to debias a model. This method could prevent potential over-fitting to a small, biased dataset. 

% used multi-lingual BERT for mitigating ethnic bias. \todo{?}
% and proposes using two monolingual BERTs for low resource language. 
   
% \vspace{-.15in}
\paragraph{Human-in-the-loop.} 
These methods involve humans to detect and mitigate biases. \citet{yao_refining_2021} used human-provided explanations to identify and reduce bias.
\citet{felkner-etal-2023-winoqueer} showed bias against marginalized communities can be mitigated using data written by that community.
\citet{chopra2020hindi} used humans to find words linking a social group to a positive or negative trait.

\paragraph{Model Unlearning}

Recently, there has been more focus on model unlearning methods (Fig. \ref{fig:lineage}). Here, the main idea is to identify and alter specific model weights that are responsible for bias.
\citet{meissner-etal-2022-debiasing} identified a subset of model weights responsible for bias and masked them during testing. The advantage of their approach is it does not require finetuning. 
\citet{lauscher_sustainable_2021, kumar-etal-2023-parameter} captured bias mitigation functionalities using ``adapters'' attached to transformer blocks. Adapters offer a unique advantage in that they can be added to the model for bias correction in a plug-and-play fashion. \citet{agarwal-etal-2023-peftdebias} improved on adapters by adjusting weights with data augmentation, then finetuning for specific tasks with fixed weights to prevent relearning.

Works in \textit{D1} offer greater ease of implementation, with customizable solutions for each model. However, as the prevalence of large language models grows, they are being trained on enormous amounts of data. In such cases, bias becomes %deeply ingrained within them models, making post-training debiasing increasingly challenging.
more difficult to mitigate after models have been trained.

\subsubsection{Debiasing during Training - \textit{D2}}
These works apply debiasing at the pre-training time or to word embeddings used at initialization.

\paragraph{Debiased word embeddings}
\citet{bolukbasi_man_2016} proposed a hard debiasing technique aimed at reducing gender bias in embeddings by adjusting the vector deviations between gendered and gender-neutral terms, offering these adjusted embeddings as an alternative to standard Word2Vec embeddings. \citet{park_reducing_2018, zhao-etal-2018-learning} further illustrate the effectiveness of debiased embeddings in reducing gender bias in LMs.

% \citet{bolukbasi_man_2016} proposed a hard debiasing technique aimed at equalizing gender associations in embeddings. Their method compensates for differences in average vector deviations between female and male gender terms relative to gender-neutral vocabulary. They published pre-trained embeddings using hard-debiasing, for use in place of Word2Vec embeddings. \citet{park_reducing_2018} and \citet{zhao-etal-2018-learning} also presented results on use of debiased word embeddings to reduce gender bias in language models.

%find the vector representation of the gender to 
% comensate for %its deviation and
% differences in deviations in average vector distances involving female versus male gender vocabulary, equalizing where possible with respect to gender-neutral  vocabulary.
%equalize some terms with respect to the neutral gender. 

\paragraph{Loss function}
Several methods employ specialized loss functions to minimize bias during model pre-training.
\citet{garimella_he_2021} used declustering loss to reduce bias. \citet{bordia_identifying_2019} proposed a loss regularization method.
\citet{huang_reducing_2020} proposed a three-step curriculum training using the distance between the embeddings as a fairness loss to reduce sentiment bias. 
\citet{liu_authors_2021} and \citet{he-etal-2022-mabel} used adversarial training and contrastive loss respectively to reduce bias in LMs.
\citet{li-etal-2023-prompt} shows that using contrastive learning during training helps in debiasing.

% aimed to reduce the significance of sociodemographic attributes in the input using adversarial training.  

% \citet{liu_authors_2021} aimed to reduce the significance of sociodemographic attributes in the input using adversarial training.  
% \cite{he-etal-2022-mabel} uses contrastive loss on counterfactual generated data to reduce gender bias.

\paragraph{Expert Models for Bias Reduction}
Recently methods using an auxiliary model, %often known as
or so-called expert model, to reduce bias have gained prominence (cf. Fig. \ref{fig:lineage}).
\citet{orgad-belinkov-2023-blind} predicted biased samples using an auxiliary model and performed sample reweighting to downweight these samples during pre-training.
\citet{jeon-etal-2023-improving} used binary classifiers, %also known as 
referred to as bias experts, to identify biased examples within a specific class.
\citet{zhang_model_2023} used gradient-based explanations to focus on sensitive attributes and downstream tasks, adjusting the training process to balance fairness and performance effectively.

% {orgad-belinkov-2023-blind} performs a sample reweighting during training by predicting biased samples using an auxiliary model that tries to predict the main model’s accuracy.

% \cite{zhang_model_2023} leverage gradient-based explanation to find two model focuses, 1) one focus for predicting sensitive attributes and 2) the other focus for predicting downstream task labels, and second, use them to perturb the latent code that guides the training of downstream task models towards fairness and utility goals. 

% \cite{gao-etal-2022-kernel} proposes a representation normalization method, where they map the embeddings to a isotropic latent space which makes the data distribution more uniform and less biased during training and helps in generalization of the model.

% \cite{richardson_add_2023} proposed a new approach using iterative training -- look into it

\subsubsection{Debiasing at Inference Time- \textit{D3}}
% \subsection{Debiasing at Inference Time- \textit{D3}} %-aaai format
% \todo[inline]{Name seems awkward, how about "Debiasing At Inference Time"}
%There are some works that focus on debiasing NLP models during inference time. These approaches can be used to reduce bias of any trained model during the testing time, without increasing training time. 
% As an alternative to debiasing during model training, 
These methods apply debiasing methods at test time. In general, these methods are quite diverse.
% A method called adversarial triggering was adopted in a few works \cite{abid2021persistent, venkit2023nationality}. 
\citet{venkit2023nationality} and \citet{abid2021persistent} applied adversarial machine learning to trigger positive associations in text generative models to reduce anti-Muslim bias and nationality bias, respectively, through prompt modifications.
\citet{majumder-etal-2023-interfair} used humans to provide feedback to balance between task performance and bias mitigation. 
\citet{qian_counterfactual_2021} performed keyword-based distillation to remove bias during inference, and to block bias acquired during training. \citet{zhao_gender_2019} addressed gender bias through averaging representations for different gender vocabulary.
% , but with little reduction in bias.
\citet{schick2021self} also presents the concept of self-debiasing, in which a model can identify and eliminate biases after generating text.

Work on debiasing during inference time faces the same issues as those in \textit{D1}. They are easy to implement but give a false impression of debiasing and do not completely remove the model bias.

\section{Limitations of Current Approaches}
\label{sec:limitations}

The works surveyed here offer valuable insights %contributions 
towards understanding bias in LMs, and demonstrating many innovative approaches and methodologies that have advanced the field. 
Alongside the commendable progress, however, a thorough analysis of the body of work on bias reveals limitations. %which we outline in this section.
%However, despite these commendable achievements, in this section we outline the key limitations of works on bias.
% Based on our survey, we outline the key limitations of works on bias in NLP. To build reliable and effective approaches, our focus centers on finding limitations to provide a checklist for future works.

\textbf{Reliability issues with bias metrics.} The robustness of existing bias metrics is questionable. %Approaches in 
Metrics introduced in works within \textit{Q1} and \textit{Q3} change significantly, %due to
given minor changes in datasets or evaluation settings \cite{antoniak_bad_2021, spliethover-etal-2022-word, du_assessing_2021, valentini-etal-2022-undesirable}. Similarly, template-based methods are highly sensitive to small modifications to words used in the templates \cite{selvam-etal-2023-tail, seshadri2022quantifying, alnegheimish-etal-2022-using}.

\textbf{Use of identical templates across bias categories.} 
Many of the works using template-based approaches \cite{an-etal-2023-sodapop, smith-etal-2022-im} use the same templates to assess diverse social biases, %which may not be effective across different social bias contexts. 
without considering whether certain template features should be specific to distinct types of bias.
This approach risks conflating bias scores across categories, suggesting a need for more tailored templates to measure specific social biases accurately. Alternatively, investigation of ways to generalize across templates to a more abstract approach, as in ~\citet{omrani_evaluating_2023}, holds promise.

\textbf{Limited scope of template-based bias measurement.} 
Template-based methods often use a restricted range of templates and target words, for example, focusing on US-based names for targets. %This narrows their scope. 
Additionally, these approaches suffer from author bias, as templates are manually designed by the authors %This author bias makes their bias scores heavily dependent on template selection 
\cite{seshadri2022quantifying, pikuliak-etal-2023-depth}. 

\textbf{Gap in translating bias metrics to real-world effects.} There is a notable disconnect between bias metrics and their implications for real-world applications. Bias metrics in \textit{Q1} have been claimed to correlate poorly with real-world biases \cite{goldfarb-tarrant_intrinsic_2021, cao-etal-2022-intrinsic}.  \citet{orgad-etal-2022-gender} argued that intrinsic and extrinsic metrics do not correlate with each other. Such observations underscore the need for improvements in metric robustness and interpretability.

\textbf{Weaknesses in finetuning approaches for debiasing.} The majority of recent works on LM debiasing focus on finetuning, valued mainly for its simplicity and adaptability. However, its effectiveness is often questionable \cite{diciccio_detection_2023}. The complexity and size of modern LLMs, which require extensive data, time, and resources to train, %mean biases are deeply embedded in them. This 
make it particularly challenging to eliminate bias through finetuning. Further, these methods
%% I merged this with the above; it didn't make sense to me to have two highly overlapping limitations
%\textbf{Debiasing is sometimes superficial.} Finetuning-based debiasing methods 
treat symptoms rather than root causes of bias, adjusting model outputs to appear less biased without actually removing bias from models \cite{gonen_lipstick_2019, tokpo-etal-2023-far}. Remarkably, some debiasing techniques can potentially increase bias \cite{mendelson_debiasing_2021}. The absence of reliable bias metrics complicates the evaluation of the effectiveness of debiasing methods. % true effectiveness. 
We recommend that future works utilize a variety of metrics to thoroughly assess debiasing results.

\textbf{Overemphasis on gender bias.} Table \ref{tab:biases} shows that about half of the literature focuses solely on gender bias. Although gender bias is a significant concern, %the overwhelming focus on it overlooks other forms of sociodemographic bias. 
other types of sociodemographic bias also deserve attention. Expanding research to cover a wider range of bias categories could provide a more comprehensive understanding of bias.

\textbf{Lack of sociotechnical understanding of bias.} %In the field of NLP, we have seen very little effort to understand %how models and corresponding bias metrics have sociotechnical impacts 
The NLP literature exhibits little attention to
the sociotechnical impacts of bias \cite{venkit2023sentiment}. Similarly, there can be incomplete consideration of the complexity of sociodemographic bias \cite{blodgett_language_2020}. %A deeper exploration of bias through 
Interdisciplinary collaborations could offer more nuanced insights and improved methodologies to measure, mitigate, prevent, and assess harms from bias. %, which can help develop better quantifying and debiasing approaches, also 
%as highlighted in Section \ref{sec:understanding}.

% \paragraph{Lack of Explicit Analysis of How  Models Can Cause Social Harm.} Works on NLP bias often %simplify its potential for causing societal harm, overlooking 
% overlook the complexity of how LM bias can impact society \cite{dev2022measures}. It is crucial to differentiate when biases might have positive or negative effects and to explore exactly how LM bias can lead to societal harm. A deeper exploration into the nature and consequences of LM bias is needed to fully grasp the implications, and guide efforts to diminish or prevent social harm. %their harmful aspects.

%\textbf{Comparison of different approaches is difficult.} %Due to the different target domains of various approaches, it is often difficult to directly compare different approaches. 
\textbf{Difficulty of comparing approaches.}
A better understanding is needed of strengths and weaknesses across approaches, given that works often focus on different domains and tasks. 
\citet{kaneko-etal-2023-comparing} compared different bias evaluation approaches without %using human labels. 
requiring the expense of human labels. We need more work in the direction of reliable and cost-effective comparison among different measurement and mitigation methods.

% \paragraph{Evaluation Beyond Standard Benchmarks}
% Reliance on standard benchmarks may not fully capture the multifaceted nature of bias. Incorporating benchmarks with user studies can provide a more holistic view of bias and its effects.

% TODO : \cite{ladhak-etal-2023-pre} studies how bias propagates in downstream tasks.

\section{Checklist}
\label{sec:checklist}
A checklist can assist future work to avoid the pitfalls of previous work and build more effective and reliable measurement and debiasing methods across more types of sociodemographic bias.  %strategies. 
We present 13 questions divided into three categories. Questions 1-6 focus on bias measurement (\textbf{QB}); questions 7-8 focus on bias mitigation (\textbf{BM}); questions 9-13 are %general questions (\textbf{GQ}), 
applicable to all works on LM bias. We hope that future work guided by these questions can help authors situate their results within the broader literature on sociodemographic bias.
%To make the most of this tool, researchers should systematically evaluate their work against each relevant question, explaining how their research addresses these aspects. Each question is formulated by analysing key limitations, ensuring they directly address critical issues in the field. 

% To effectively utilize this checklist, researchers should systematically assess their work against each relevant question and answer how their work performs for each question. Each question is decided by analysing the major limitations in each of these areas. 
% These questions will help researchers avoid the pitfalls of previous works and build more effective and reliable strategies. 
% Future works should systematically assess their work against each relevant question. 

 % We do not intend any one work to address all questions; rather, we believe work that addresses multiple questions will have a significant impact.

% Questions for bias measurements
\textbf{[Q1:QB]} \textit{Robustness:} Is your bias measurement stable against small modifications to templates/descriptors?
    
\textbf{[Q2:QB]} \textit{Country-focused data:} Does your method rely on country-specific data, such as the U.S.? If so, how can it be adapted to others?

\textbf{[Q3:QB]} \textit{Real-World Relevance:} How do your bias measurements reflect real-world biases, and affect end-users?

\textbf{[Q4:QB]} \textit{Future Usability:} Have you taken measures to make sure your approach is easily extendable to ensure that it is useable after 5 years?  

\textbf{[Q5:QB]} \textit{Data Diversity:} Have you used diverse data sources to diminish biases present in the data?

\textbf{[Q6:QB]} \textit{Verification of Bias Type}: What measures have you taken to ensure your bias measurement on a given type of bias doesn't overlap or confuse with other biases?

% to ensure that bias metric for that type is not inadvertently confounded with other types of bias? 
% How do you ensure your focus on a specific bias type doesn't overlap or confuse with other biases?

%Questions for bias mitigations

\textbf{[Q7: BM]} \textit{Scalability and Efficiency:} Can your debiasing method efficiently scale to large models and datasets while maintaining effectiveness?
% Is your bias mitigation approach scalable to large models and datasets? Does it balance the trade-off between computational efficiency and the thoroughness of bias mitigation?

\textbf{[Q8: BM]} \textit{Monitoring and Evaluation:} Is there a way for you to continuously assess and adjust the effectiveness of your approach?

%Questions common to all works

\textbf{[Q9: GQ]} \textit{Extensibility to other Social Groups:} Can your method be extended to additional sociodemographic groups?

\textbf{[Q10: GQ]} \textit{Risk of Misinterpretation:} Can there be a situation when your approach falsely indicates reduced bias in models?

% \textbf{[Q11]} \textit{Potential for Misuse:} Could your method be misused, leading to long-term societal or ethical issues?

\textbf{[Q11: GQ]} \textit{Cultural Sensitivity:}  Does your approach take into account the contextual and cultural variations in language use?

\textbf{[Q12: GQ]} \textit{Interdisciplinary Insights:} Does your method integrate knowledge from multiple disciplines to understand bias?

\textbf{[Q13: GQ]}  \textit{Transparency and Reproducibility:} Can others reproduce your method and results?

% \textbf{[Q14: GQ]} \textit{Community Engagement:} Does your method allow for user and community feedback?% for continuous improvement?

\section{Future Directions}
\label{sec:future}
Looking ahead, we anticipate greater emphasis on bias mitigation at training time. Post-training bias mitigation %acts as a cure, rather than prevention of the problem and can only go to an extent. 
adds to the costliness of very large LMs, and serves as a filter rather than a corrective. Subsequent to the first drafts of this survey,
we have already %are already encouraged to 
seen progress in this direction \cite{jeon-etal-2023-improving}. %particularly as latest 
% The latest approaches using bias experts have been proven more effective than post-training corrections \cite{orgad-belinkov-2023-blind}. This is particularly relevant for large language models that require extensive compute and data resources to train. Post-training bias mitigation %acts as a cure, rather than prevention of the problem and can only go to an extent. 
% \textcolor{red}{adds to the costliness of large LMs, and serves as a filter rather than a corrective.}
The recent progress in the usage of contrastive learning during training \cite{li-etal-2023-prompt} and using expert models during training \cite{orgad-belinkov-2023-blind}, has shown to generate less biased models
and we expect more research in these directions.

Despite their growing popularity, template-based methods for measuring bias face challenges \cite{selvam-etal-2023-tail, seshadri2022quantifying}. We believe that these challenges can be tackled with careful consideration of the limitations, such as lack of robustness, leading to more effective and reliable bias measurement. %strategies.
We anticipate that prompt-based methods will gain prominence. Additionally, integrating interdisciplinary insights with algorithmic analysis will likely gain traction for quantifying and mitigating sociodemographic bias.

Finally, as robust methodologies emerge, we anticipate increased hope for and emphasis on intersectional bias, the overlap of multiple types of bias.

% We feel with advent of robust approaches, going forward there will be more focus on intersectional bias, intersection of multiple biases, in LMs.

\section{Conclusion}

We have presented a comprehensive literature survey %encompassing 273 relevant
based on the iterative consideration of 273 works on sociodemographic bias in LMs. Our proposed typology provides an overview of the current research landscape. We identified promising directions for future research and introduced a 13-question checklist designed to guide future research towards more effective and reliable approaches and to avoid pitfalls of previous works. We encourage increased reliance on interdisciplinary methods to better capture and address the nuances of sociodemographic bias in LMs.

\section{Limitations}
In our survey, we focused on works from ACL Anthology, NeurIPS proceedings, FAccT, and AIES. We might have missed some relevant works in our survey, that appeared in other venues.
While we have systematically organized the bias literature into categories as shown in Fig. \ref{fig:main}, which came from an extensive survey of current literature, our framework might not encompass all existing or future research. We would like to emphasize that most of the works covered in this survey focus on the English language and some approaches discussed might not transfer to other languages. Additionally, our emphasis on sociodemographic bias means that valuable insights from works addressing other forms of bias in language models were not covered in our analysis.

\section{Bias Statement}
In this work, we provide a comprehensive survey of works on sociodemographic bias in language models. We defined sociodemographic bias as the difference in model performance across social groups. 
Such bias has the potential for harm in a real-world setting. Our definition applies to prominent demographic distinctions such as gender identity 
(male, female, non-binary), or income-based groupings (e.g., low, middle, and high income), or other broad-coverage distinctions that are learnable by LMs. For example, associating ``Caucasian man'' with ``handsome'', and "African-American man" with ``angry'' is a clear indication of bias in models \cite{garimella_he_2021}. In occupation-related tasks, associating ``receptionist'' with ``she'', and  ``philosopher'' with ``he'' can have harmful effects in real-world settings \cite{bolukbasi_man_2016}.

\section{Ethics Statement}

Our work addresses the ethical impact of sociodemographic bias in NLP, offering a comprehensive survey of 273 peer-reviewed articles to highlight the presence and implications of bias within language models. By systematically organizing research findings and tracking bias approaches over the past decade, our work promotes transparency, awareness, and accountability within and beyond the NLP community. The survey provides a meticulously designed checklist, based on the weaknesses and limitations of the field, to guide future research toward more effective solutions for mitigating bias. 

We also emphasize the social and ethical implications of bias underscoring the significance of addressing these issues to prevent potential negative consequences. We hope that our analysis aids in shaping more inclusive and equitable NLP technologies by fostering dialogue, awareness, and proactive measures to address sociodemographic bias, incorporating ideas beyond the field of NLP.

% \input{content/08_02_ethics_statement}


\bibliography{references}
\bibliographystyle{acl_natbib}

\appendix

\section{Appendix}
\label{sec:appendix}

\subsection{Evaluation Datasets}
\label{sec:datasets}

% \todo[inline]{For this section, "word association" would be more informative than "context association"} 
% \todo[inline, color=yellow]{Double check on task for each dataset.}
% \todo[inline, color=green]{Rewrote the section as per prior discussion. Double-check the writing style and content.}

% There are many different datasets which has been published which aims at evaluating bias in NLP models. Twitter hate speech detection dataset <cite and elaborate> is used in many works \cite{subramanian_evaluating_2021}<cite more> to measure bias. Occupation classification dataset is also used to measure bias on occupation dataset. \cite{bartl_unmasking_2020} provides a dataset of 2700 sentence pairs to capture gender bias with respect to professions. \cite{webster_measuring_2021} performs swapping to generate 16,980 pairs to measure gender bias using Semantic Textual Similarity. \cite{zhao_gender_2018} proposed WinoBias dataset. \cite{nadeem_stereoset_2020} provides a dataset of 2 tasks - fill in the blanks and choose appropriate sentence. Bias is measured as likelihood of attribute sentence following the context sentence.

% Datasets on bias in NLP are a crucial resource for researchers and practitioners seeking to develop more fair and equitable natural language processing systems. These datasets typically consist of examples of biased language or text, as well as annotations or labels indicating the type of bias present.

Bias benchmark datasets provide valuable resources for NLP fairness research. 
% Datasets on NLP bias are critical resources for researchers and practitioners seeking to develop more equitable NLP systems. 
These datasets commonly contain illustrative examples of biased language, often templated sentences filled with contrastive social group terms. Datasets allow standardized bias evaluation on diverse tasks using controlled examples. 
% This section reviews datasets created to assess sociodemographic bias, where each dataset consists of templates filled with terms and names representing different social groups.  
Many of them focus on a particular type of language context, such as co-reference, sentiment, or question answering, while others probe for stereotype bias through word associations. 
% One of the largest consists of prompts for text generation. 
% Table~\ref{tab:datasets} %%%SUBMISSION MODIFICATION
Table present in the \textit{Appendix} summarizes these datasets.

% In the realm of bias identification in NLP, %natural language models, 
% one of the predominant approaches is to employ datasets capable of detecting sociodemographic bias.
% % in a particular NLP architecture.
% To facilitate understanding of these datasets, %we provide an overview of the most common publicly available bias measurement datasets in NLP. 
% % Our study focuses on research papers that explicitly aim to publish datasets that anyone can use to identify a specific form of bias in various NLP applications. 
% % These datasets are accessible to the public and widely available.
% we present a catalog of ten public datasets intended for identifying bias. Each dataset comprises templates consisting of naturally occurring sentences used to assess a model's performance concerning bias against a particular group. These identification or evaluation datasets fall into one of the following five categories based on the type of application they focus on: \textit{coreference resolution, context association, question answering, sentiment evaluation,} and \textit{text generation}.

%For instance, 
In the case of \textit{coreference resolution}, \citet{zhao_gender_2018} proposed a method for identifying gender bias using Winograd-schema sentences %with entities denoting people referred to by their profession. 
for occupation terms.
\citet{webster2018mind} introduced GAP, a gender-balanced, labeled corpus of 8,908 ambiguous pronoun–name pairs designed to detect gender bias in coreference resolution. In the %\textit{context association} 
\textit{word association}
domain, \citet{nangia_crows-pairs_2020} presented CrowS-Pairs, a sentence pair corpus that measures a model's bias by assessing if it favors sentences with stereotypes. %on a non-stereotype sentence. 
\citet{nadeem_stereoset_2020} released StereoSet, a large-scale natural dataset in English designed to measure stereotypical bias using inter- and intra-sentence association of words to stereotypical contexts. \citet{li_unqovering_2020} proposed UNQOVER, a general framework for probing bias in \textit{question answering} models %by constructing questions to see if a model associates a group with its corresponding stereotype.
using questions to probe whether a model associates a sociodemographic group to a stereotype. \citet{smith-etal-2022-im} published HolisticBias, consisting of 450,000 unique sentence prompts for measuring 13 types of sociodemographic bias in generative LMs.

In the domain of \textit{sentiment evaluation}, 
% these corpora contain sentences tailored to capture sociodemographic bias in sentiment and toxicity analysis models. 
\citet{kiritchenko2018examining} released EEC, an 8,640 English sentence collection %carefully 
curated to test bias toward certain races and genders in sentiment analysis models. 
BITS \cite{venkit2021identification, venkit2023automated} is a similar corpus of 1,126 sentences
%Similarly, \citet{venkit2021identification} released the BITS corpus, comprising 1,126 sentences 
curated to measure disability, race, and gender bias in sentiment and toxicity analysis models. %In the context of \textit{text generation} applications, \citet{smith-etal-2022-im} published HolisticBias, consisting of 450,000 unique sentence prompts used to measure 13 sociodemographic bias in generative sentence models.

%Datasets on NLP bias are critical resources for researchers and practitioners seeking to develop more equitable and unbiased natural language processing systems. These datasets typically comprise examples of biased language or text and annotations or labels indicating the type of bias present. We comprehensively summarize these datasets in our \textit{Appendix}.

% More datasets - \cite{felkner-etal-2023-winoqueer}

% We tried to categorise works in one of the following categories based on their main contribution. Sometimes, works have major contribution in multiple categories and maybe present multiple times below. Thus total number of works mentioned below are more than 214, but the total number of unique works is 214.%\footnote{A github link will be provided upon acceptance.} 
Table \ref{tab:datasets} provides list of datasets for quantifying bias in NLP models.

 \begin{table*}
\small
\centering
\begin{tabular}{c|c|c|c}
\toprule 
            % \multicolumn{4}{c}{MCAN}     \\ \hline
% Dataset name                    &       Task                    &   Bias Type   &   Train size    &   Val size      &   Test Size \\ \hline
Dataset name                    &       Task                    &   Bias Type     &   Dataset Size \\ \hline
% Twitter hate speech             &       Hate Speech             &               &   2000          &   1000          &   1500    \\
\makecell{WinoBias \\ \cite{zhao_gender_2018}} &      \makecell{Coreference Resolution}  &   \makecell{Gender}     % &   1,580          &     -  
&   1,580    \\

\makecell{WinoGender \\ \cite{rudinger_gender_2018}}                      &       \makecell{Coreference Resolution}                       &    \makecell{Gender}      &  %  -    &   -     &  
720     \\

\makecell{GAP \\ \cite{webster2018mind}}                      &      \makecell{Coreference  Resolution}                       &    \makecell{Gender}      &  % -    &   -     & 
8,908     \\

\makecell{Counter-GAP \\ \cite{xie2023counter}}  & \makecell{Coreference Resolution}                     &   Gender      &    % -  &   -    &  
4,008    \\

\makecell{CrowS-Pairs \\ \cite{nangia_crows-pairs_2020}}  &     \makecell{Word Association}             &   
\makecell{Gender, race, religion, \\ age, sexual  orientation, \\nationality, disability,\\physical appearance, \\and socioeco. status.}   &   %  -      &   -      &      
1,508      \\   
\makecell{StereoSet \\ \cite{nadeem_stereoset_2020}}  & \makecell{Word Association}                     &   \makecell{Race, gender, \\ religion, and profession}      &    % -  &   -    &  
16,995    \\

\makecell{WikiGenderBias \\ \cite{gaut_towards_2020}}                      &      \makecell{Word Association}                       &    \makecell{Gender}      &  % -    &   -     &  
45,000     \\
\makecell{UnQOVER \\ \cite{li_unqovering_2020}}                      &      \makecell{Word Association}                       &    \makecell{Gender, Nationality,\\Ethnicity,Religion}      & %  -    &   -     &  
8,908     \\
\makecell{WinoGrande \\ \cite{sakaguchi_winogrande_2021}}                      &       \makecell{Word Association}                       &    \makecell{Dataset\\Bias}     &   %40,938    &   1,267     &
1,767     \\

\makecell{BBQ \\ \cite{parrish2021bbq}}                      &       \makecell{Word Association}                       &    \makecell{9 Sociodemographic Group}      &  %  -    &   -     &  
58,492     \\

\makecell{EEC \\ \cite{kiritchenko2018examining}}                      &       \makecell{Sentiment Evaluation}                      &    \makecell{Gender, Race}     &  % -    &   -     &  
8,640     \\
\makecell{BITS \\ \cite{venkit2021identification}}                      &      \makecell{Sentiment Evaluation}                       &    \makecell{Gender, Race,\\ Disability}      &  % -    &   -     &  
1,126     \\
\makecell{HolisticBias \\ \cite{smith-etal-2022-im}}                      &       \makecell{Text Generation}                       &    \makecell{13 Sociodemographic Group}      &  %  -    &   -     &  
450,000     \\

\bottomrule 
\end{tabular}

\caption{List of Evaluation datasets used to measure bias in NLP models}
\vspace{-1.5em}
\label{tab:datasets}

\end{table*}

\subsection{List of papers surveyed}
Below is the list of papers surveyed in this work, sorted based on our taxonomy.

\paragraph{Explicit Bias(T1)} : 

% Bias : 
\cite{katelyn_bias_2023, deas-etal-2023-evaluation, liu_authors_2021, de-arteaga_bias_2019, bell_simplicity_2023, silva-etal-2021-towards, park_reducing_2018, sap-etal-2020-social, b-etal-2021-overview, lauscher-glavas-2019-consistently, rozado_wide_2020, rudinger-etal-2017-social, shah_predictive_2020, du-etal-2022-understanding, nozza-etal-2022-pipelines, honnavalli-etal-2022-towards, lucy-bamman-2021-gender, mendelson_debiasing_2021, matthews-etal-2021-gender, cao-etal-2022-intrinsic, Papakyriakopoulos_2020_bias, kementchedjhieva_john_2021, garrido-munoz_survey_2021, strengers_2020_adhering, delobelle-etal-2022-measuring, Fisher2020, sheng-etal-2020-towards, zhang-etal-2020-demographics, hendricks2018women, mehrabi_survey_2021, mayfield_equity_2019, schwartz2021proposal, nozza_2019_unintended, vaidya2019empirical, he-etal-2019-unlearn, hovy-sogaard-2015-tagging, wolfe_low_2021, sakaguchi_winogrande_2021, agarwal-etal-2019-word, white-cotterell-2021-examining, luo-glass-2023-logic}

Gender Bias : \cite{sharma-etal-2022-sensitive, kaneko-etal-2022-gender-bias, stahl-etal-2022-prefer, kaneko-etal-2023-comparing, toro-isaza-etal-2023-fairy, hada-etal-2023-fifty, attanasio-etal-2023-tale, goldfarb-tarrant-etal-2023-cross, lee-etal-2023-target, gaut_towards_2020, sun_mitigating_2019, hamidi_gender_2018, zhou_examining_2019, savoldi-etal-2021-gender, sahlgren-olsson-2019-gender, ahn-etal-2022-knowledge, tal-etal-2022-fewer, kaneko-etal-2022-gender, field-tsvetkov-2020-unsupervised, garimella-etal-2019-womens, escude-font-costa-jussa-2019-equalizing, bhaskaran-bhallamudi-2019-good, mccurdy2020grammatical, kaneko-bollegala-2019-gender, larson-2017-gender, du_assessing_2021, bartl_unmasking_2020, webster_measuring_2021, tan_assessing_2019, bolukbasi_man_2016, maudslay_its_2020, zhao_gender_2019, rudinger_gender_2018, lu_gender_2019}

Racial Bias : \cite{goldfarb-tarrant-etal-2023-cross, levy-etal-2023-comparing, field_examining_2023, cheng_how_2023, sap_risk_2019, hanna_towards_2020, blodgett-etal-2016-demographic, davidson_racial_2019, friedman-etal-2019-relating, shen2018darling, karve-etal-2019-conceptor, nadeem_stereoset_2020, garimella_he_2021, nangia_crows-pairs_2020, tan_assessing_2019, wei_detecting_2021, brown_language_2020}

Disability bias : \cite{venkit2021identification, venkit2022study, hutchinson_social_2020, bennett_what_2019, whittaker2019disability, hassan-etal-2021-unpacking-interdependent, narayanan2023towards}

Ethnicity bias : \cite{bauer-etal-2023-social, levy-etal-2023-comparing, malik-etal-2022-socially, li-etal-2022-herb, ahn_mitigating_2021, garg_word_2018, li_unqovering_2020, abid2021persistent, manzini-etal-2019-black, venkit2023nationality, bhatt_re-contextualizing_2022}, Nationality bias - \cite{ladhak-etal-2023-pre, levy-etal-2023-comparing, venkit2023unmasking}, Political bias - \cite{thebault_diverse_2023, shen2018darling, rozado_wide_2020}, Age bias \cite{nangia_crows-pairs_2020, diaz_addressing_2018} and sexual-orientation bias \cite{anaelia_im_2023, nangia_crows-pairs_2020, cao_toward_2020}

% \paragraph{Implicit Bias(T2)} : \cite{white-cotterell-2021-examining, agarwal-etal-2019-word, provilkov_multi-sentence_2021, venkit2022study, karimi-mahabadi-etal-2020-end, de-arteaga_bias_2019, kementchedjhieva_john_2021}

\paragraph{Distance based metrics(Q1)} : \cite{caliskan_semantics_2017, dev2019attenuating, zhao-etal-2017-men, basta-etal-2019-evaluating, shen2018darling, pmlr-v97-brunet19a, may_measuring_2019, dev_oscar_2021, zhou_examining_2019, pujari2019debiasing, sutton2018biased, lauscher_general_2020, wei_detecting_2021, bolukbasi_man_2016, ross-etal-2021-measuring, tan_assessing_2019, ethayarajh_understanding_2019, chaloner_measuring_2019, bordia_identifying_2019, valentini-etal-2023-interpretability}

\paragraph{Probing based metrics(Q2)} : \cite{orgad-etal-2022-gender,  immer-etal-2022-probing, chen2021probing, limisiewicz-marecek-2021-introducing, kennedy-etal-2020-contextualizing, sweeney-najafian-2019-transparent, tan-etal-2020-morphin, mendelson_debiasing_2021, white-etal-2021-non}

\paragraph{Performance metrics(Q3)} : \cite{de-arteaga_bias_2019, kwon-mihindukulasooriya-2022-empirical, zhang_interpreting_2022, huang_reducing_2020, dixon_measuring_2018, zhao_gender_2018, cho-etal-2019-measuring, stanovsky_evaluating_2019, gonen-webster-2020-automatically, borkan_nuanced_2019, dev_measuring_2019}

\paragraph{Prompt based metrics(Q4)} : \cite{nagireddy2023socialstigmaqa, webster_measuring_2021, smith-etal-2022-im, kurita_measuring_2019, krishna-etal-2022-measuring, bhaskaran-bhallamudi-2019-good, gupta_2022_CVPR, prabhakaran_perturbation_2019, ahn_mitigating_2021, bartl_unmasking_2020, li_unqovering_2020, venkit2021identification, salazar-etal-2020-masked,  dev_measuring_2019,diaz_addressing_2018, zhang_2020_hurtful, garg_counterfactual_2019, helm_evaluation_2022, kusner_counterfactual_2018, huang_reducing_2020, akyurek-etal-2022-measuring, gardner-etal-2020-evaluating, ousidhoum-etal-2021-probing, parrish-etal-2022-bbq, kiritchenko2018examining, touileb-etal-2023-measuring, gupta2023calm, pikuliak-etal-2023-depth, touileb-etal-2023-measuring, an-etal-2023-sodapop, felkner-etal-2023-winoqueer, ousidhoum-etal-2021-probing}

\paragraph{Debiasing during Finetuning(D1)} : \cite{ungless-etal-2022-robust, du-etal-2023-towards, omrani-etal-2023-social, zhou-etal-2023-causal, thakur-etal-2023-language, jin-etal-2021-transferability, he_controlling_2022, zmigrod_counterfactual_2019, jin-etal-2021-transferability, gaci-etal-2022-debiasing, gupta-etal-2022-mitigating, ghaddar-etal-2021-context, kumar-etal-2020-nurse, han-etal-2021-diverse, attanasio-etal-2022-entropy, joniak-aizawa-2022-gender, chopra2020hindi, maudslay_its_2020, park_reducing_2018, yao_refining_2021, liang_towards_2020, sen-etal-2022-counterfactually, ma-etal-2020-powertransformer, limisiewicz-marecek-2022-dont, yang-etal-2021-simple, wang-etal-2021-eliminating-sentiment, pujari2019debiasing, sedoc-ungar-2019-role, tan-etal-2020-morphin, sutton2018biased, ravfogel-etal-2020-null, kaneko-bollegala-2019-gender, karve-etal-2019-conceptor, gyamfi2020deb2viz, shin-etal-2020-neutralizing, zhang-etal-2020-demographics, wen-etal-2022-autocad, chopra2020hindi, yang2020causal, lu_gender_2019, lauscher_sustainable_2021, garg_counterfactual_2019, dev_measuring_2019, dev_oscar_2021, manzini-etal-2019-black, bolukbasi_man_2016, ahn_mitigating_2021,orgad-etal-2022-gender, felkner-etal-2023-winoqueer, de-vassimon-manela-etal-2021-stereotype}

\paragraph{Debiasing during Training (D2)} : \cite{an-etal-2022-learning, bolukbasi_man_2016, he-etal-2019-unlearn, han-etal-2022-balancing, liu-etal-2020-hyponli, escude-font-costa-jussa-2019-equalizing, prost-etal-2019-debiasing, james2019probabilistic, park_reducing_2018, zhao-etal-2018-learning, gao-etal-2022-kernel, Sweeney_2020_reducing, hube_2020_debiasing, sen2020towards, saunders-byrne-2020-reducing, dixon_measuring_2018, karimi-mahabadi-etal-2020-end, he-etal-2022-mabel, richardson_add_2023}
% karimi-mahabadi-etal-2020-end -> might want to add details
Loss functions for bias mitigation : \cite{hashimoto2018fairness, qian-etal-2019-reducing, berg-etal-2022-prompt, romanov-etal-2019-whats, garimella_he_2021, bordia_identifying_2019, huang_reducing_2020, provilkov_multi-sentence_2021, liu_authors_2021, orgad-belinkov-2023-blind, li-etal-2023-prompt}

\paragraph{Debiasing during Inference (D3)} : \cite{majumder-etal-2023-interfair, qian_counterfactual_2021, zhao_gender_2019, abid2021persistent, guo-etal-2022-auto, schick-etal-2021-self, venkit2023nationality}

% Other literature surveys : \cite{shah_predictive_2020, garrido-munoz_survey_2021, delobelle-etal-2022-measuring, czarnowska_quantifying_2021}

% \todo{Find a way to order the appendix better}
% \todo{Put at the end. These works are difficult to categorize}
\paragraph{Works on Bias}: These are works that are difficult to categorize in one of the above categories.  
\cite{ Chouldechova_2020_snapshot, green2019good, Zhang2018EqualityOO, mayfield_equity_2019, katell_toward_2020, dwork_fairness_2011, jacobs2020meaning,  k_towards_2022, czarnowska_quantifying_2021, blodgett-etal-2021-stereotyping, zhuo2023exploring, mulligan2019thing, jacobs_measurement_2021, schoch-etal-2020-problem, franklin_2022_ontology, bender2019typology, espana-bonet-barron-cedeno-2022-undesired, hutchinson_50_2019, bender_2021_on, goldfarb-tarrant_intrinsic_2021, brown_language_2020, li_unqovering_2020, bagdasaryan2019differential, liu-etal-2020-gender, zhiltsova2019mitigation, chopra2020hindi, luo2023perspectival, shah_predictive_2020, garrido-munoz_survey_2021, delobelle-etal-2022-measuring, czarnowska_quantifying_2021}

% \section{Datasets}

\end{document}